\RequirePackage{amsmath}
\documentclass[runningheads]{llncs}

\usepackage{graphicx}
\usepackage{xcolor}
\usepackage{xspace}

\usepackage{subcaption}
\usepackage{siunitx}

\usepackage[colorlinks=true,allcolors=blue,urlcolor=blue]{hyperref} 

\providecommand{\etal}{\emph{et al.\,\xspace}}
\providecommand{\ie}{\emph{i.e.\,\xspace}}
\providecommand{\eg}{\emph{e.g.\,\xspace}}

\providecommand{\EQ}[1]{Eq.~({#1})}
\providecommand{\FIG}[1]{Figure~{#1}}
\providecommand{\FIG}[2]{Figures~{#1} and~{#2}}
\providecommand{\SEC}[1]{Section~{#1}}

\providecommand{\VECTOR}[1]{\mathbf{#1}}
\providecommand{\MATRIX}[1]{\mathbf{#1}}

\newcommand{\abs}[1]{\left\lvert#1\right\rvert}
\newcommand{\norm}[1]{\left\lVert#1\right\rVert}
\newcommand{\transpose}{^\top}
\newcommand{\RR}{\bbbr}

\DeclareMathOperator*{\argmax}{arg\,max}
\DeclareMathOperator*{\argmin}{arg\,min}

\begin{document}
%
%
\title{Intrinsic Calibration of Depth Cameras for Mobile Robots using a Radial Laser Scanner}
%
\titlerunning{Intrinsic Calibration of Depth Cameras for Mobile Robots}
\author{David Zu\~{n}iga-No\"{e}l \and Jose-Raul Ruiz-Sarmiento \and Javier Gonzalez-Jimenez}
\authorrunning{D. Zu\~niga-No\"el \etal}
%
\institute{Machine Perception and Intelligent Robotics group (MAPIR). Dept. of System Engineering and Automation. Instituto de Investigaci\'on Biom\'edica de Malaga (IBIMA). University of Malaga. Spain.
\\
\email{\{dzuniga,jotaraul,javiergonzalez\}@uma.es}}
\maketitle              
%


\begin{abstract}
Depth cameras, typically in RGB-D configurations, are common devices in mobile robotic platforms given their appealing features: high frequency and resolution, low price and power requirements, among others.
These sensors may come with significant, non-linear errors in the depth measurements that jeopardize robot tasks, like free-space detection, environment reconstruction or visual robot-human interaction. 
This paper presents a method to calibrate such systematic errors with the help of a second, more precise range sensor, in our case a radial laser scanner. In contrast to what it may seem at first, this does not mean a serious limitation in practice since these two sensors are often mounted jointly in many mobile robotic platforms, as they complement well each other. Moreover, the laser scanner can be used just for the calibration process and get rid of it after that. The main contributions of the paper are: i) the calibration is formulated from a probabilistic perspective through a Maximum Likelihood Estimation problem, and ii) the proposed method can be easily executed automatically by mobile robotic platforms.
To validate the proposed approach we evaluated for both, local distortion of 3D planar reconstructions and global shifts in the measurements, obtaining considerably more accurate results. A C++ open-source implementation of the presented method has been released for the benefit of the community.

\end{abstract}


\section{Introduction}
\label{sec:introduction}


Nowadays, many mobile robots get awareness of their workspaces using  RGB-D cameras~\cite{jing2017comparison,raul2017ijrr}. These compact and affordable sensors provide per-pixel depth measurements along with colour information at high frame rates, simplifying a variety of robotic tasks that would be more involved if using a regular camera only, such as 3D object detection and localization~\cite{schwarz2018object,ruiz2017survey}, safe autonomous navigation~\cite{jaimez2015nav}, or map building/scene reconstruction~\cite{infinitam,jamiruddin2018rgbd,raul2017kbs}, among others.
Alternative sensors providing 3D depth information are LiDAR~\cite{zhang2014loam} or Time-of-Flight cameras~\cite{Foix2011tof}, but they are not as widely spread as structured-light depth sensors, mainly due to their higher price~\cite{Rusu20113d}.

Unfortunately, affordability of structured-light depth cameras comes at a cost: depth estimates are affected by significant distortion, not always well modeled by factory calibration parameters~\cite{song2018rgbd,fiedler2013kinect}. These errors can be unacceptable for some common robotic applications, and thus require a further calibration by the user. For example, we empirically observed that an obstacle-free path through an open door can be narrowed by intrinsic depth errors up to a point where it appears to the robot as a colliding path. We also experienced the negative effect of inaccurate measurements in algorithms for plane segmentation, scene reconstruction, and human pose estimation~\cite{raul2017ijrr,sarmiento2013kinect,fdzmoral2014calib}.


With the massive deployment of robotic platforms~\cite{dellacorte2019ral}, calibration methods suitable to be executed automatically by robots are desirable, seeking to prevent the manual calibration of each depth sensor prior to deployment.
However, existing intrinsic calibration methods for structured-light depth cameras cannot be easily automated. For example, the method described in~\cite{Teichman2013} aims to correct depth measurements via visual SLAM and, therefore, has the underlying requirement of a well illuminated, textured enough environment.
Authors in~\cite{Cicco2015nonparametric} argue that their method could be executed automatically, however, it is applicable only for sensors mounted with a near zero pitch angle. Recently, authors in~\cite{Basso2018} proposed another calibration approach based on the observation of a known checkerboard pattern with a regular RGB camera. In order to enable automatic calibration, their approach requires manipulating the environment to include the visual pattern which, in turn, hampers the deployment process.

In this paper, we first empirically analyze the behaviour of structured-light depth cameras and then present a method to compensate for systematic errors in the measurements, which can be easily executed automatically by mobile robotic platforms.
More precisely, the proposed method requires observing, at different distances, a vertical planar surface (\eg a wall) from both the depth camera and another extrinsically calibrated sensor (\eg a 2D laser scanner, device commonly found in robotic platforms) not suffering from those errors. In this way, the second sensor is used to obtain depth references for calibration. Note that planar surfaces are ubiquitous in human-made environments and specific visual calibration patterns are not required.
Bias functions for intrinsic depth errors are then calibrated in a Maximum Likelihood Estimation framework.
The output of the calibration method are per-pixel quadratic bias functions from which systematic errors in depth measurements can be corrected in an online fashion.

To demonstrate the suitability of our proposal, we collected data from two RGB-D cameras and a 2D laser scanner mounted on a mobile robot (the robotic platform Giraff~\cite{gonzalez2012technical}) when approaching a vertical, planar surface, and carried out an experimental evaluation showing both quantitative and qualitative performance results. A C++, ROS integrated open-source implementation of the presented method is available at:
\url{https://github.com/dzunigan/depth_calibration}


\section{Related Work}
\label{sec:related_work}

Early works in depth error calibration aimed to calibrate distortions along with the extrinsic parameters with respect to an RGB camera. For example, the authors in~\cite{Zhang2011depth} considered the calibration of an RGB-D camera pair resorting to a linear depth distortion function, while Herrera \etal~\cite{Herrera2012joint} tackled the calibration of two colour cameras and a depth one. In the latter case the disparity distortion was modelled as a per-pixel offset with exponential decay governed by two global parameters. 
Both approaches employ planar surfaces for depth compensation, tendency that still holds in recent works. An example of this is the work by Basso \etal~\cite{Basso2018}, which proposed a calibration method based on the observation of a planar pattern with a regular camera, while the extrinsic calibration is more a ``side effect''. 

All above-mentioned works require a visual pattern (typically a checkerboard) in order to compute reference depth measurements, and must be included in the robot workspace to enable automatic calibration. There are works not requiring this, like~\cite{Cicco2015nonparametric}, where the authors proposed a non-parametric calibration approach and they get rid of the visual pattern requirement. However, to perform the calibration on a mobile robot, another sensor (\eg laser scanner) is required in order to provide reference values, and only sensors mounted with a near zero pitch angle can be calibrated.

Another way to get rid of known visual patterns is by using a visual SLAM pipeline to provide the depth references. To the best of our knowledge, depth correction via SLAM was first introduced by Teichman \etal~\cite{Teichman2013}. Their method makes the strong assumption that the errors at close ranges (below \SI{2}{\meter}) are negligible, and thus are used as reference within the SLAM pipeline. Depth correction factors are then estimated for each pixel and at a number of fixed distances. Another work based on a similar idea was presented in~\cite{Quenzel2017depth}, where the authors assume known extrinsic calibration between the RGB and the depth cameras. Their method projects features from a sparse map (generated from the RGB camera) into the depth camera poses in order to estimate the correction factors. They use the thin plate spline as a tool for approximating a dense representation of the sparse correction factors. The main drawback of these methods is that they have the underlying requirements of well illuminated and textured enough environments in order to provide reliable estimates.

The calibration method presented in this work does not require any visual pattern and thus can be easily executed automatically by mobile robots. As in~\cite{Cicco2015nonparametric}, another sensor is needed to provide reference measurements, concretely a radial laser scanner. Notice that this assumption is not very restrictive, since these sensors are commonly used in mobile robotic platforms. Moreover, we can use the laser scanner temporally just for the calibration process and get rid of it after that. In contrast to~\cite{Cicco2015nonparametric}, we argue that the systematic depth errors can be well modeled from a more compact parametric representation. Additionally, our method does not assume a specific orientation of the depth camera to carry out the calibration.

\section{Depth Error Model}
\label{sec:error_model}

\begin{figure}[!t]
    \centering
    \includegraphics[width=0.9\textwidth]{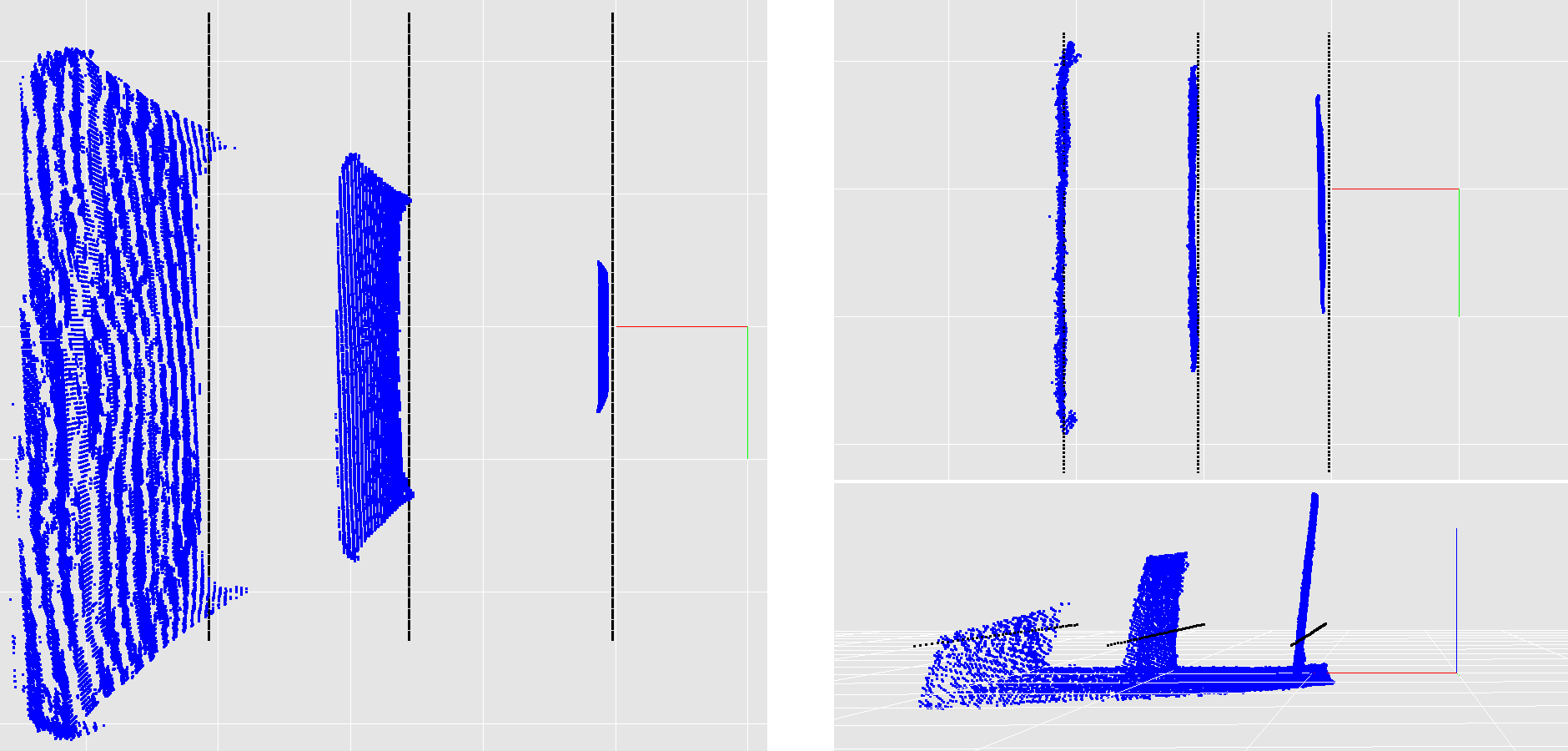}
    \caption{Illustration of the errors and their variation with distance. Left, a depth camera observing a perpendicular wall at \num{1}-\SI{4}{\meter}. Right, another camera with a \SI{60}{\deg} pitch observing the same wall, at \num{1}--\SI{3}{\meter}. Note that the reconstructed ground is parallel to the $x$-$y$ plane, while the wall has a noticeable inclination.}
    \label{fig:distortion}
\end{figure}

In this work, as in~\cite{Basso2018}, we consider both the ``local distortion'' and ``global'' errors as the main source of systematic errors. The \emph{local distortion} has the characteristic effect of deforming the resulting point cloud, while the \emph{global errors} shift the average observed depth. Illustrative examples of these errors are shown in \FIG{\ref{fig:distortion}}. We argue that both sources of error can be explained by a depth bias $\beta_{u,v}$:
\begin{equation}
    z_{u,v} = z_{u,v}^* + \beta_{u,v},
\end{equation}
for each pixel $(u, v) \in \Omega$ in the image domain independently, where $z_{u,v}^*$ and $z_{u,v}$ represent the true and the measured depths, respectively. We consider the bias to be normally distributed:
\begin{equation}
    \beta_{u, v} \sim \mathcal{N}\big(\mu_{u, v}(z_{u,v}), \sigma^2(z_{u,v})\big),
\end{equation}
where $\mu_{u, v}$ is a per-pixel mean function and $\sigma$ is a global standard deviation function modeling the uncertainty in the measurements.

\begin{figure}[t]
    \centering
    \begin{subfigure}[b]{0.45\textwidth}
        \includegraphics[width=\textwidth]{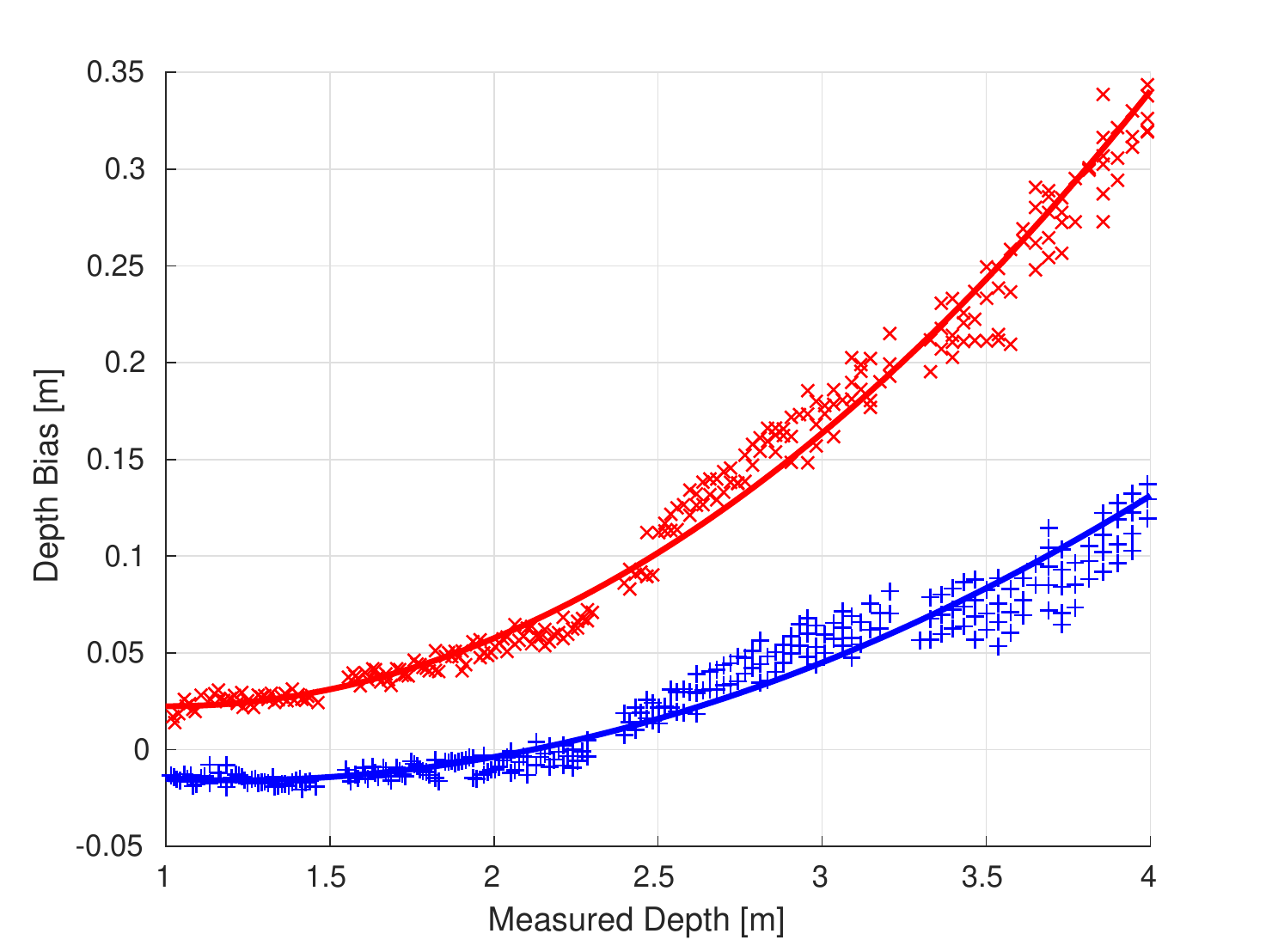}
        \caption{}
        \label{fig:depth_bias}
    \end{subfigure}
    \begin{subfigure}[b]{0.45\textwidth}
        \includegraphics[width=\textwidth]{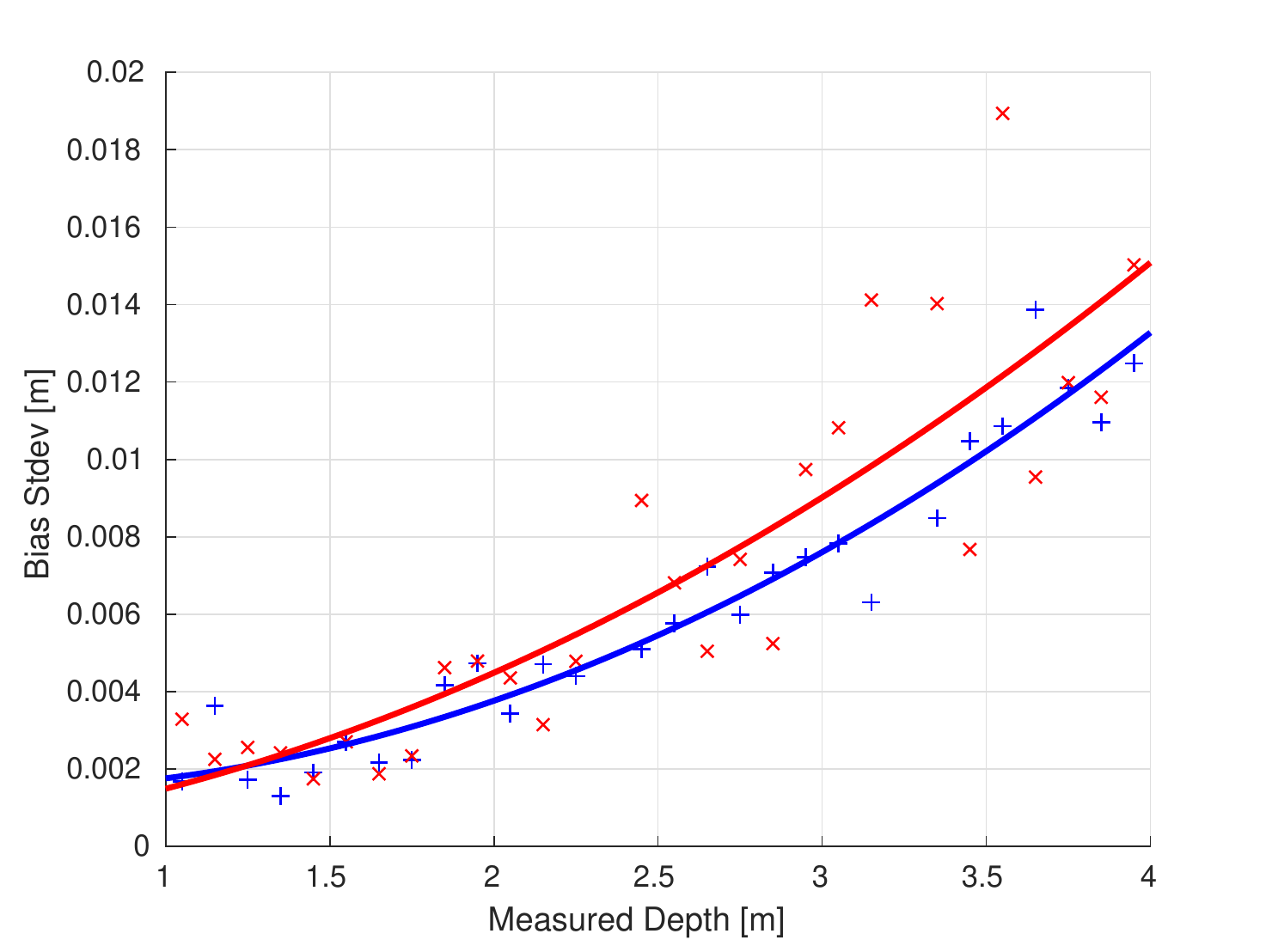}
        \caption{}
        \label{fig:bias_noise}
    \end{subfigure}
    \caption{The observed bias (\ref{fig:depth_bias}) and bias noise (\ref{fig:bias_noise}) as a function of the measured depth, along with quadratic curve fits, for two different pixels.}
\end{figure}

The bias, computed as the difference between the measured depth and the real one, are plotted in \FIG{\ref{fig:depth_bias}} as a function of the measured depth, for different pixels. The lines in that figure represent fitted quadratic models (see \SEC{\ref{sec:calibration_mle}}). It becomes clear that each pixel is affected by a different bias, but the evolution of the biases with respect to depth are well explained by quadratic functions.

Regarding the uncertainty in the measurements, previous research~\cite{Smisek2011kinect} found that it follows a quadratic evolution with respect to depth. We verified this behaviour empirically by analyzing the standard deviation of the measurements in a similar setting as for the biases. The standard deviation plotted against the measured depth are reported in \FIG{\ref{fig:bias_noise}}. Notice that, unlike the bias, the uncertainty of the measurements is similar for different pixels. This phenomenon has also been considered in our framework by modeling a single variance function for all pixels (see \SEC{\ref{sec:calibration_mle}}). 

Finally, the systematic depth errors can be compensated by subtracting the bias mean:
\begin{equation} \label{eq:depth_compensation}
    \bar{z}_{u,v} = z_{u,v} - \mu_{u,v}(z_{u,v}) = z_{u,v}^* + \epsilon, \quad \epsilon \sim \mathcal{N}\big(0, \sigma^2(z_{u,v})\big),
\end{equation}
yielding unbiased depth measurements.

\section{Calibration Approach}
\label{sec:calibration}

In this section we describe the proposed calibration approach. First, \SEC{\ref{sec:calibration_depth_reference}} discusses the process of computing depth reference measurements from the observation of a planar surface by the sensors. Then, the formulation of the calibration problem in a Maximum Likelihood framework and its solution are described in \SEC{\ref{sec:calibration_mle}}.

\subsection{Computation of the Depth References}
\label{sec:calibration_depth_reference}

The input of the calibration method are observations of a vertical, planar surface from both a depth camera and another sensor not suffering from the same errors. For the former, observations are in the form of depth images, while for the latter they are in the form of geometric parameters of the observed plane. These parameters are $(\VECTOR{n}, d) \in \RR^3 \times \RR^+$ such that:
\begin{equation}\label{eq:plane}
    \VECTOR{n} \cdot \VECTOR{x} - d = 0,
\end{equation}
for any point $\VECTOR{x} \in \RR^3$ lying on the plane. Here, $\VECTOR{n}$ represents the unit normal vector (from the origin to the plane) and $d \geq 0$ the perpendicular distance to the origin (Hessian normal form).

The extrinsic calibration $(\MATRIX{R}, \VECTOR{t}) \in \text{SE(3)}$ between the two sensors allow us to express the plane parameters observed by the second sensor into the coordinate system of the depth camera. Clearly, the new normal vector $\VECTOR{n}'$ is affected only by the rotation $\MATRIX{R}$, while the new distance $d'$ can be computed as:
\begin{equation}
    \VECTOR{n}' = \MATRIX{R} \VECTOR{n}, \quad d' = -\VECTOR{n}' \cdot \VECTOR{t} - d,
\end{equation}
which is the distance of the new origin $-\VECTOR{t}$ from the rotated coordinates (before translation).

Depth cameras allow to reconstruct 3D points via back-projection, using the associated depth measurements and the intrinsic camera parameters (provided by the manufacturer). We parameterize the 3D line representing an incoming ray with respect to depth $z \in \RR$ as:
\begin{equation}
    \VECTOR{l}_{u,v}(z) = z \Big( \frac{u - c_x}{f_x}, \frac{v - c_y}{f_y}, 1 \Big)\transpose,
\end{equation}
where $(c_x, c_y) \in \RR^2$ refers to the camera center, and $f_x, f_y \in \RR$ to the focal lengths (in each axis). In this way, the reconstructed 3D point can be computed as $\VECTOR{l}_{u,v}(z_{u,v})$. Therefore, we define the reference depth measure $z_{u,v}^*$ such that:
\begin{equation}\label{eq:true_depth}
    \VECTOR{n}' \cdot \VECTOR{l}_{u,v}(z_{u,v}^*) - d' = 0,
\end{equation}
\ie enforcing the plane constraint in \EQ{\ref{eq:plane}} on the reconstructed 3D point. Finally, since \EQ{\ref{eq:true_depth}} is linear with respect to $z_{u,v}^*$, the solution can be computed as:
\begin{equation}
    z_{u,v}^* = \frac{d'}{\VECTOR{n}' \cdot \VECTOR{l}_{u,v}(1)},
\end{equation}
yielding pairs $(z_{u,v},z_{u,v}^*)$ that relate measured and reference depth values.


\subsection{Maximum Likelihood Estimation of the Bias Functions}
\label{sec:calibration_mle}

Once having computed the depth measurement-reference pairs, the estimation of the bias functions is divided into two main steps. First, we fit a quadratic function to the observed deviations, which is common for all pixels (recall \FIG{\ref{fig:bias_noise}}). Next, for each pixel independently, we solve for the actual bias parameters (recall \FIG{\ref{fig:depth_bias}}).

In first place, we want to estimate the parameters of a quadratic function that best represents the evolution of the bias noise. In a Least Squares sense, this is:
\begin{equation}
    \argmin_{a,b,c} \sum_{k \in \Pi} \norm{\sigma_k - \sigma(k)}^2, \quad \sigma(k) = ak^2 + bk + c,
    \label{eq:ls_optimization}
\end{equation}
given the discrete deviation samples $\sigma_k$ over the discrete sampling interval $\Pi$. In order to compute observed standard deviations, we divide the observed bias into discrete bins for each pixel independently:
\begin{equation}
    S^k_{u,v} = \{ z - z^* \mid t > \abs{z - k}, \forall (z, z^*) \in M_{u,v} \},
\end{equation}
where $t \in \RR$ is a discretization threshold and $M_{u,v}$ is the set of depth pairs for a pixel $(u,v) \in \Omega$. Then, for each set of observations $S_k$ with $k \in \Pi$, we compute the deviation $\sigma_k$ as: \vspace{-0.1cm}
\begin{equation} \label{eq:discrete_deviation}
    \sigma^2_k = \frac{1}{\sum_{(u,v) \in \Omega} \abs{S^k_{u,v}}} \sum_{(u,v) \in \Omega} \Big( \sum_{z \in S^k_{u,v}} (z - \bar{S}^k_{u,v})^2 \Big),
\end{equation}
where $\abs{S}$ represents the set's cardinality and $\bar{S}$, the mean. Equation~\ref{eq:discrete_deviation} aims to compute the variance of a range of depth measurements, where each pixel can have a different bias mean. This way, we obtain the discrete samples $\sigma_k$ used to fit the function modeling the bias noise in \EQ{\ref{eq:ls_optimization}}.


Having an estimation of the uncertainty, we proceed to solve for the parameters of a quadratic approximation to the bias function for each pixel independently. We formulate the calibration problem in a Maximum Likelihood Estimation fashion as:\footnote{Hereafter, we drop the $u,v$ subscript to improve readability.} \vspace{-0.3cm}
\begin{equation}\label{eq:mle_bias} 
    \argmax_{a,b,c} \prod_{i=1}^N p\big(z_i \mid z_i^*, \mu(z_i), \sigma(z_i)\big), \quad \mu(z) = az^2 + bz + c,
\end{equation}
for a likelihood function $p$ and $N$ independent observations. Under the assumption of normality, the likelihood function becomes:
\begin{equation}
    p\big(z \mid z^*, \mu, \sigma\big) = \frac{1}{\sqrt{2\pi\sigma^2}}\exp\Big(-\frac{(z - z^* - \mu)^2}{2\sigma^2} \Big).
\end{equation}

Taking the negative logarithm of \EQ{\ref{eq:mle_bias}} yields an equivalent Least Squares problem: \vspace{-0.1cm}
\begin{equation}\label{eq:ls_bias}
    \argmin_{a,b,c} \sum_{i=1}^N \frac{1}{\sigma^2(z_i)} \norm{z_i - z_i^* - \mu(z_i)}^2,
\end{equation}
which has a closed form solution since the residual expression is linear with respect to the optimization parameters. In this way, solving \EQ{\ref{eq:ls_bias}} we approximate a bias function that can be used to compensate for the measured depths in a per-pixel fashion.

\section{Experimental Evaluation}
\label{sec:evaluation}

The goal of the experimental evaluation is to validate our approach in a real setting.
In this respect, we provide a quantitative evaluation of how well the error model presented in \SEC{\ref{sec:error_model}} can handle both the local distortions (\SEC{\ref{sec:evaluation_local_distortion}}) and the global errors (\SEC{\ref{sec:evaluation_global_error}}). We also show qualitative improvements in 3D reconstructions after calibration (\SEC{\ref{sec:qualitative_evaluation}}).

\begin{figure}[t]
    \centering
    \includegraphics[width=0.9\textwidth]{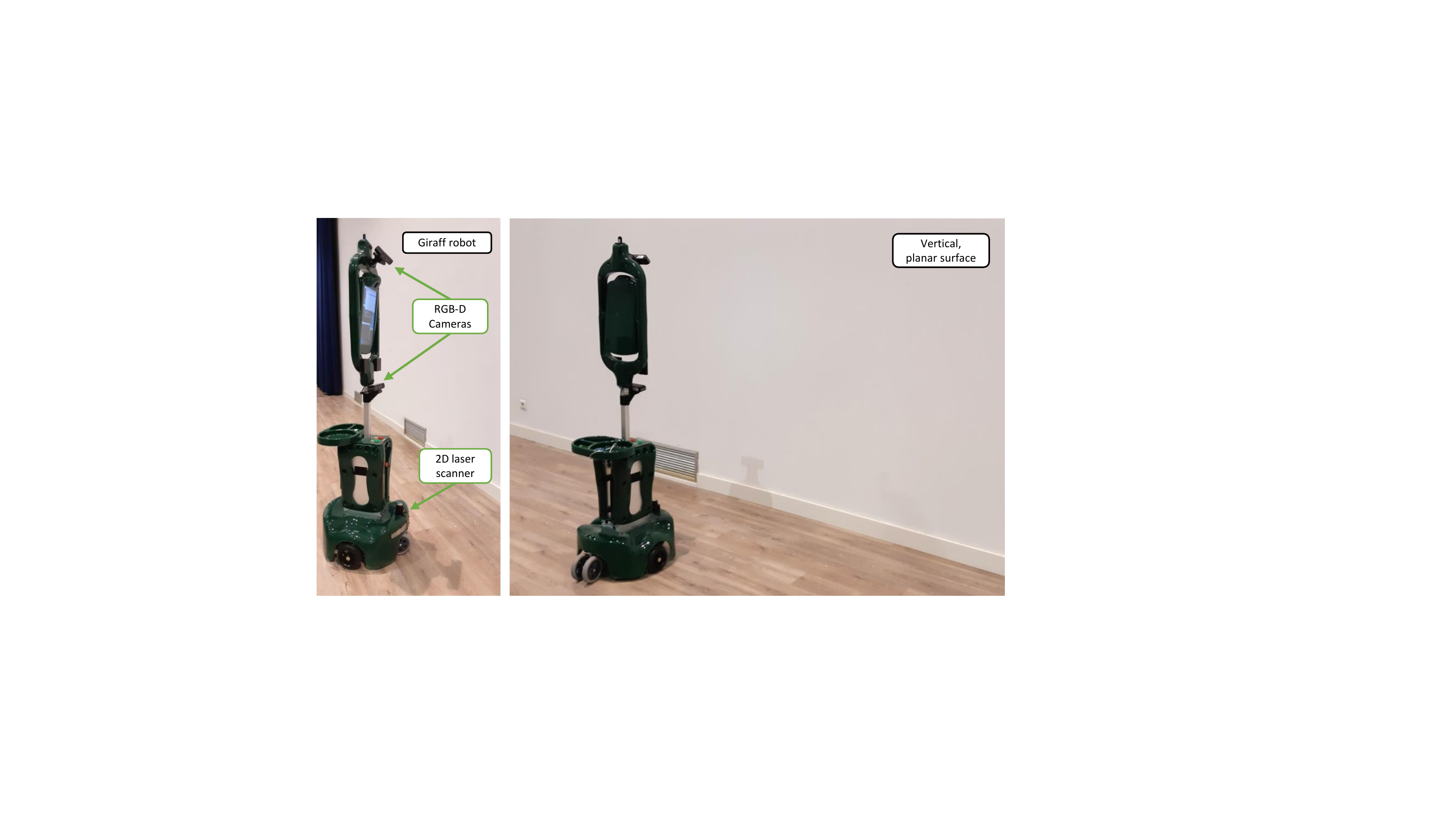}
    \caption{Left, Giraff robot with annotations of the sensors involved in the calibration process. Right, the robot facing a planar surface during data collection.}
    \label{fig:robot}
\end{figure}

To carry out these evaluations, we recorded two independent sequences with two RGB-D sensors (Orbbec Astra) and a 2D laser scanner (Hokuyo URG-04LX-UG01). The sensors were mounted on a Giraff~\cite{gonzalez2012technical} mobile robot and we recorded the sequences while moving it towards and away a wall.
The sensor setup and a snapshot of the collection procedure are depicted in \FIG{\ref{fig:robot}}. Note that the upper RGB-D camera has a non-negligible pitch, while the other camera and the laser are mounted horizontally, \ie with near zero pitch. The extrinsic calibration parameters between the sensors were estimated using the automatic multi-sensor method proposed in~\cite{zuniga2019automatic}.
From the two recorded sequences, one was used to perform the depth calibration described in \SEC{\ref{sec:calibration}}, while the other one was used for evaluation purposes. For the sake of reproducibility, the collected data is available at: \url{https://doi.org/10.5281/zenodo.2636878}.

\subsection{Local Distortion Evaluation}
\label{sec:evaluation_local_distortion}

In order to evaluate the undistortion performance, we follow a similar approach as described in~\cite{Basso2018}. Since the local distortion errors deform the reconstructed 3D structure, the evaluation method consists of fitting a plane to the point cloud acquired while observing a wall, and then computing the Root Mean Square (RMS) perpendicular distance to the extracted plane for each point belonging to the planar surface.

This is, for a plane $\pi$, we have:
\begin{equation} \label{eq:perpendicular_error}
    e_{\bot}(\pi) = \sqrt{\frac{1}{N} \sum_{i=1}^N \norm{\VECTOR{n}_\pi \cdot \VECTOR{x}_i - d_\pi}^2},
\end{equation}
for each 3D point $\VECTOR{x}_i \in \RR^3$ of the planar surface.

\begin{figure}[t]
    \centering
    \begin{subfigure}[b]{0.48\textwidth}
        \includegraphics[width=\textwidth]{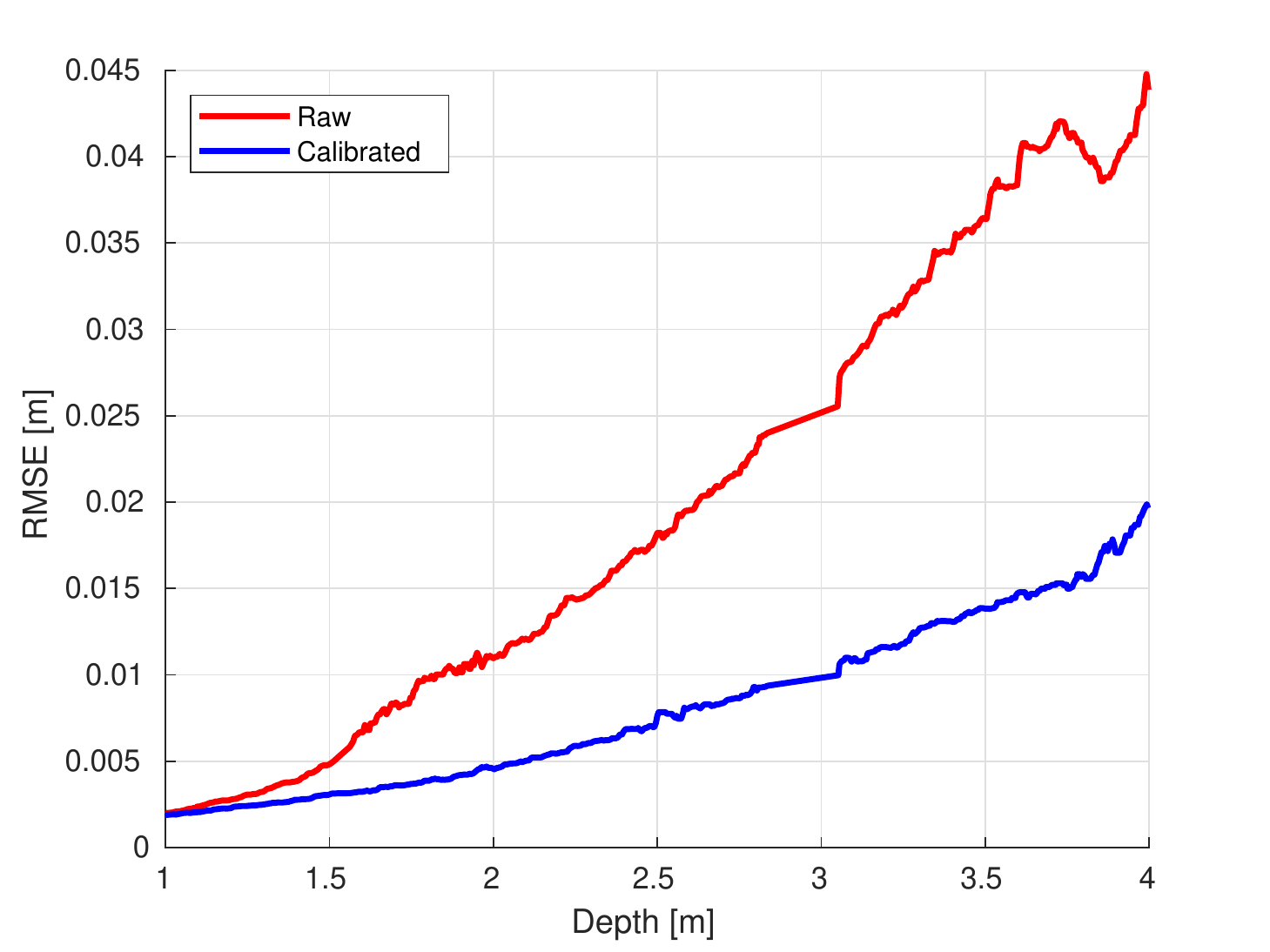}
        \caption{}
        \label{fig:local_cam0}
    \end{subfigure}
    \begin{subfigure}[b]{0.48\textwidth}
        \includegraphics[width=\textwidth]{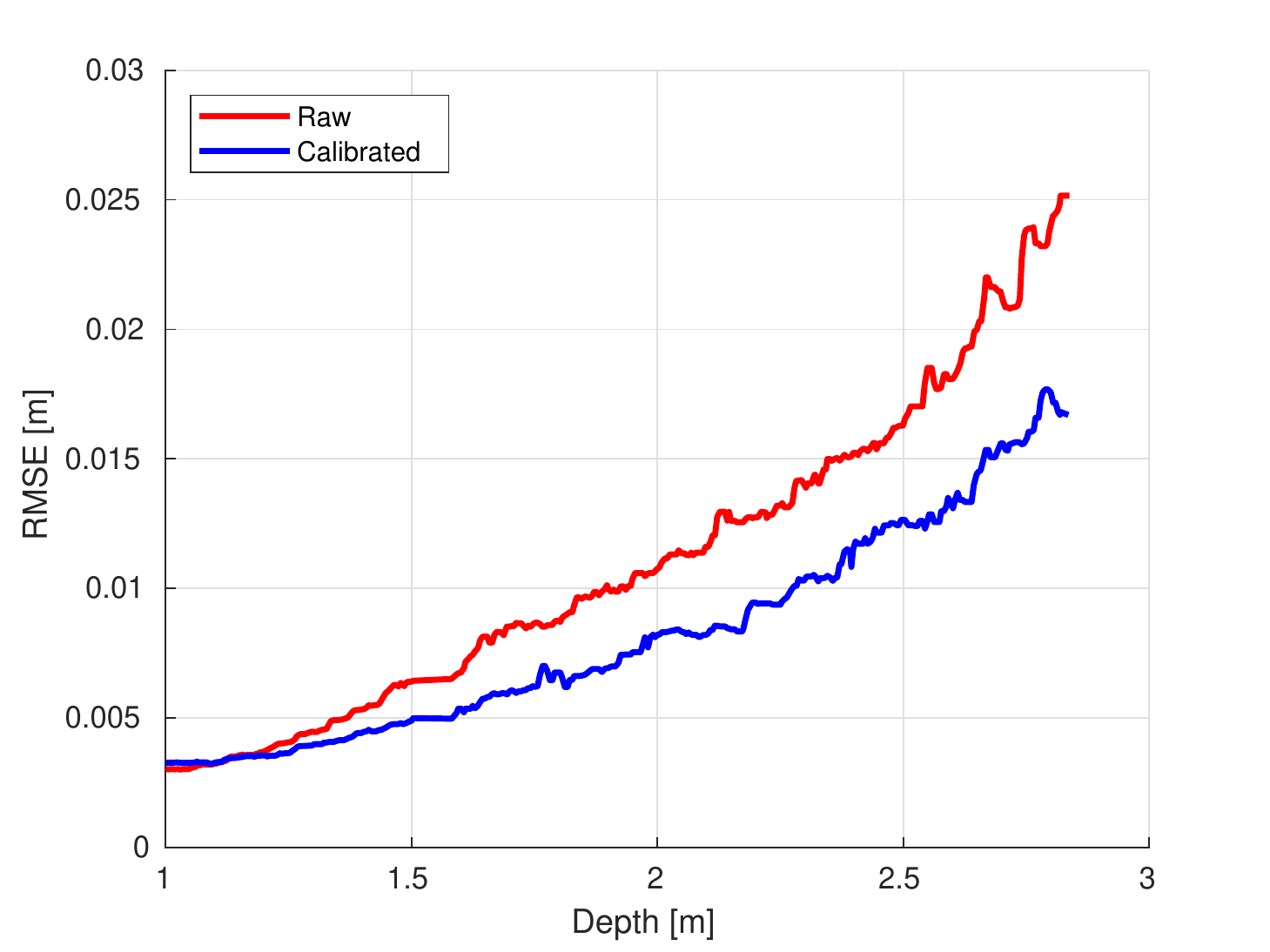}
        \caption{}
        \label{fig:local_cam1}
    \end{subfigure}
    \caption{Local distortion performance evaluation for the two RGB-D cameras (\ref{fig:local_cam0}~lower camera; \ref{fig:local_cam1}~upper camera). In both cases, calibrated depth measurements show better performance.}
\end{figure}

The evaluation results for the lower and upper cameras, in terms of the RMS perpendicular error, are shown in \FIG{\ref{fig:local_cam0}} and~\ref{fig:local_cam1}, respectively.
We can see that calibrated depth measurements achieve better performance in both cases when compared to the original ones. For example, the calibration improves $\sim$\SI{2.5}{\centi\meter} the RMSE at \SI{4}{\meter} for the lower camera, and $\sim$\SI{1}{\centi\meter} near \SI{3}{\meter} for the upper one.
It is also noticeable the difficulties that calibrated measurements have in reaching error-free measurements, and how the error grows with respect to depth. This behaviour can be explained by the quantization error of the sensor, as argued in~\cite{Basso2018}. In the case of the upper camera, errors are even larger. This phenomenon is caused by the nonzero pitch angle of the camera, as noise in the measurements increases when observing surfaces away from the perpendicular orientation, as shown in~\cite{Nguyen2012kinect_noise}. 

\subsection{Global Error Evaluation}
\label{sec:evaluation_global_error}

In order to evaluate the global error, we follow a similar approach as before, but in this case we computed the perpendicular error with respect to a reference plane. Recall that the global error shifts the measurements away from their true value. Thus, we can evaluate the error function in \EQ{\ref{eq:perpendicular_error}} with respect to the plane as observed by the laser scanner.

\begin{figure}[t]
    \centering
    \begin{subfigure}[b]{0.48\textwidth}
        \includegraphics[width=\textwidth]{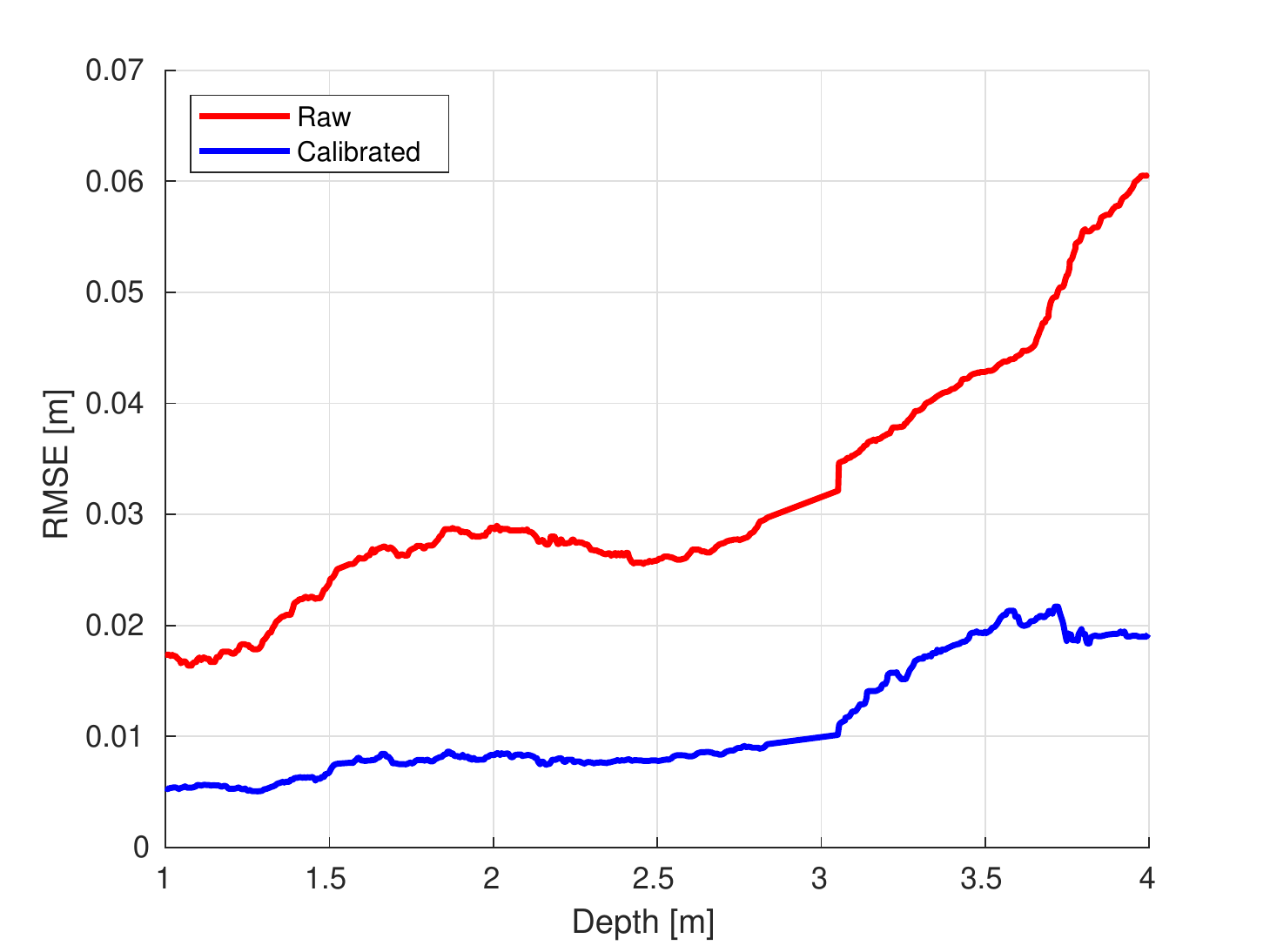}
        \caption{}
        \label{fig:global_cam0}
    \end{subfigure}
    \begin{subfigure}[b]{0.48\textwidth}
        \includegraphics[width=\textwidth]{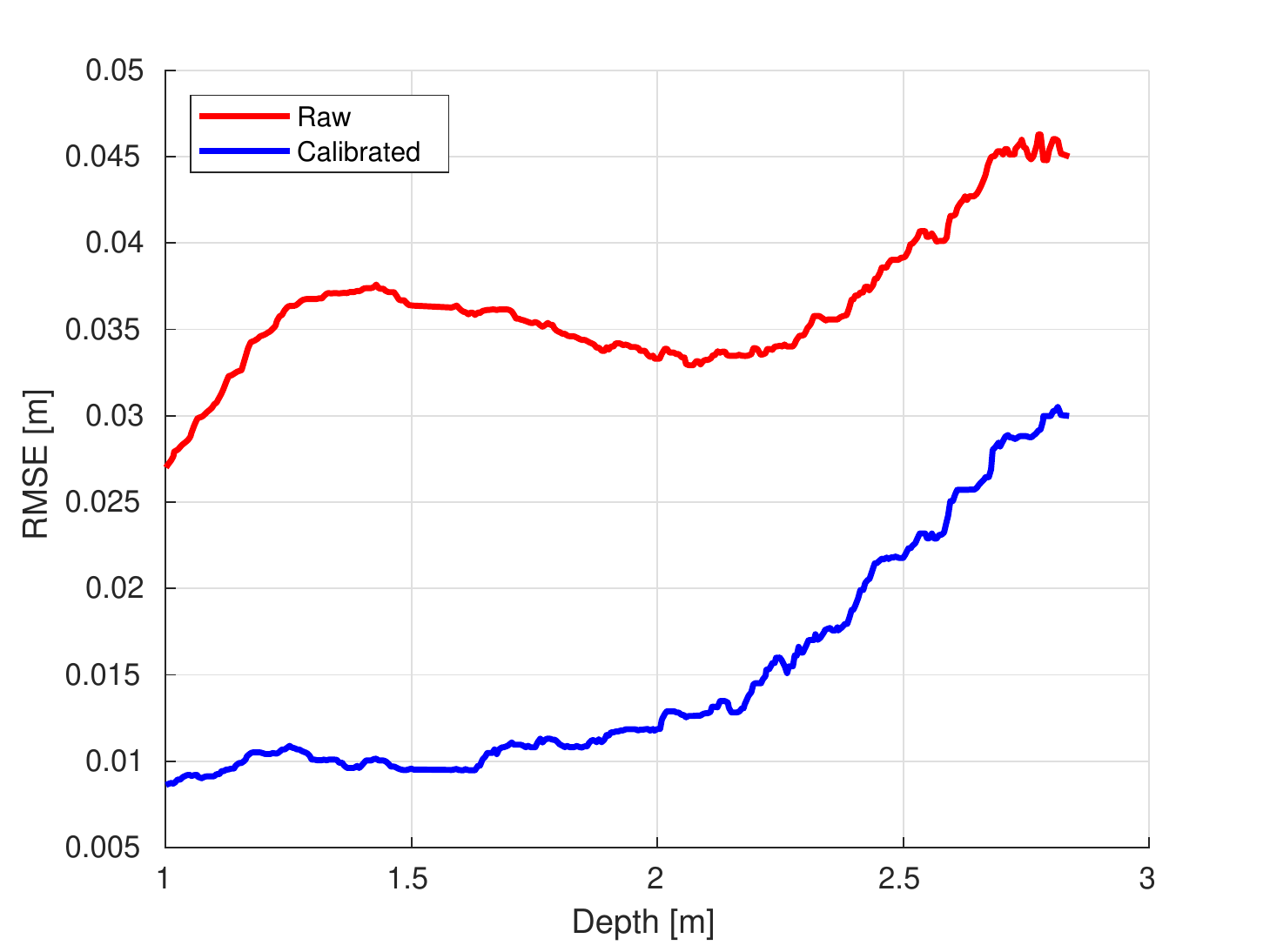}
        \caption{}
        \label{fig:global_cam1}
    \end{subfigure}
    \caption{Global distortion performance evaluation for the two RGB-D cameras (\ref{fig:global_cam0}~lower camera; \ref{fig:global_cam1}~upper camera). A significantly lower error is shown when using calibrated measurements.}
\end{figure}

The RMSE of the compensated and original depth measurements are reported in Figures~\ref{fig:global_cam0} and~\ref{fig:global_cam1} for the lower and upper cameras, respectively. 
Here, the calibration improves up to \SI{4}{\centi\meter} the RMSE for the lower camera (at \SI{4}{\meter}) and up to \SI{2.5}{\centi\meter} for the upper one (at \SI{1.5}{\meter}).
Errors also tend to grow with depth, for the same reasons as before. Additionally, global errors after calibration are higher than the local distortion ones. This is mainly due to other external sources of errors affecting the evaluation, as \eg errors in the laser measurements, extrinsic calibration errors or time delays between the laser and the cameras. 
Despite of this, in both cases, the use of calibrated depth measurements improves the accuracy of the measurements. 

\subsection{Qualitative Evaluation}
\label{sec:qualitative_evaluation}

In this section, we provide a qualitative evaluation of the obtained, reconstructed 3D point clouds when using compensated depth measurements compared to the original  ones. For that purpose, we compare the reconstruction of vertical walls to ground truth measurements before and after calibration. For space reasons, only the results for the lower camera are shown.

\begin{figure}[t]
    \centering
    \includegraphics[width=0.9\textwidth]{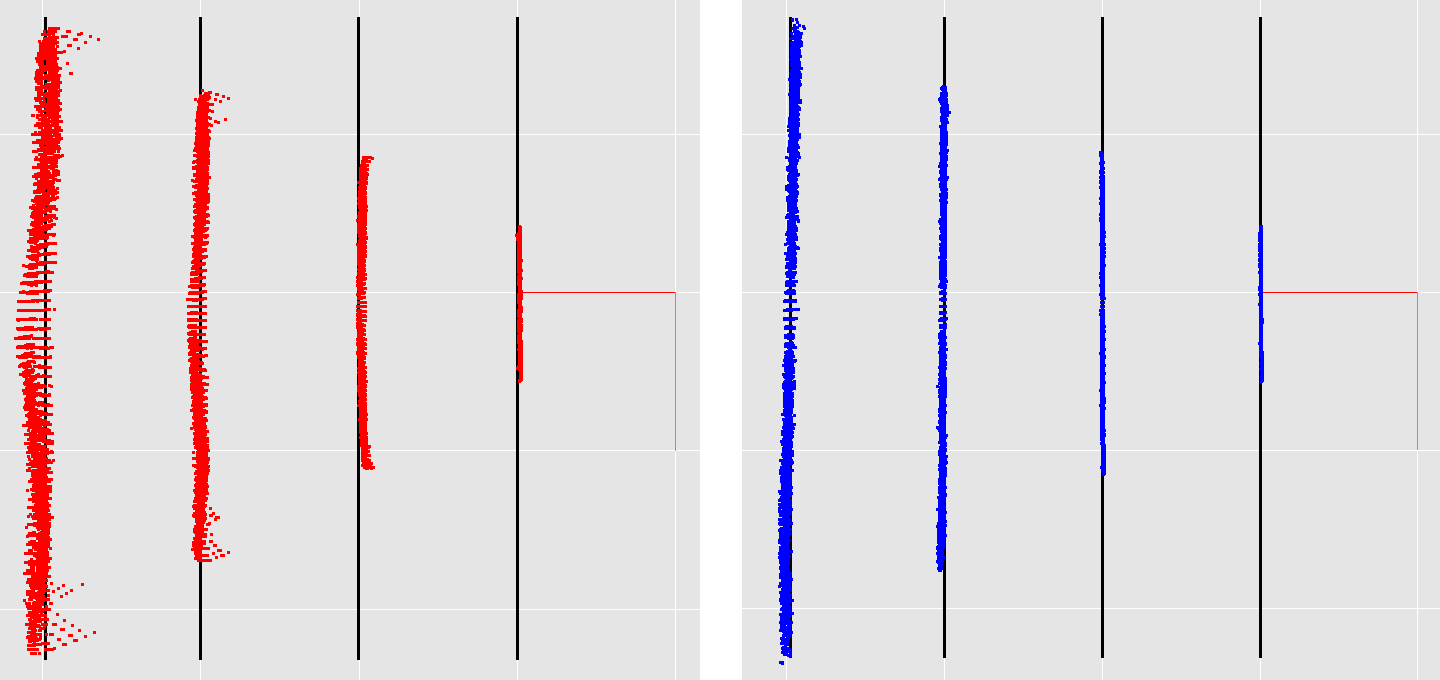}
    \caption{Reconstructed point clouds form raw (left) and calibrated measurements (right), with reference measurements shown as black lines at \num{1}--\num{4}~\si{\meter}.}
    \label{fig:qualitative_cam0}
\end{figure}

The reconstructed point clouds using the raw, original depth measurements are shown in \FIG{\ref{fig:qualitative_cam0}}-left, while \FIG{\ref{fig:qualitative_cam0}}-right reports the corrected point clouds after calibration. At closer distances, the distortion is negligible, while small offsets are noticeable in the raw measurements. At higher distances, distortions are clearly visible. It can be observed that, after calibration, both the small offsets in the measurements as well as distortions are significantly corrected.

\section{Conclusion}
\label{sec:conclusion}

In this work, we presented a method to calibrate systematic errors arising from depth cameras that can be easily executed by mobile robot platforms. First, we analyzed and characterized these errors, and then proposed a calibration method based on Maximum Likelihood Estimation. 
This method requires to observe a planar surface with both the depth camera and another sensor (\eg a radial laser scanner), used to compute reference depth measurements. The output of the calibration are per-pixel parametric bias functions that can be used to compensate for these systematic depth errors.
We evaluated the proposed method in a real robotic platform equipped with two RGB-D cameras and a 2D laser scanner, and showed that the proposed model can handle both local distortions and global errors, producing considerably more accurate measurements. We also provided a qualitative evaluation of the method performance, reporting noticeable error corrections. 
In the future we plan to incorporate a robust plane detection mechanism in order to enhance the method performance in cluttered environments.  

\vspace{0.3cm}

\begin{small}
\noindent \textbf{Acknowledgments.} This work has been supported by the research projects \emph{WISER} (DPI2017-84827-R), funded by the Spanish Government and the European Regional Development's Funds (FEDER), \emph{MoveCare} (ICT-26-2016b-GA-732158), funded by the European H2020 program, the European Social Found through the Youth Employment Initiative for the promotion of young researchers, and a postdoc contract from the {I-PPIT} program of the University of Malaga.
\end{small}

%
%
%
%
\bibliographystyle{splncs04}
\bibliography{biblio.bib}

\end{document}